\documentclass[10pt,letterpaper]{article}

\usepackage{cogsci}

\cogscifinalcopy 

\usepackage{pslatex}
\usepackage{apacite}
\usepackage{float}

\usepackage{times}
\usepackage{latexsym}
\usepackage{color}
\usepackage{graphicx}
\usepackage[utf8]{inputenc}
\usepackage[russian,english]{babel}
\usepackage{mathastext}
\usepackage{amsmath}
\usepackage{amssymb}
\usepackage{comment}
\usepackage{paralist}
\usepackage{ctable}
\usepackage{xspace}
\usepackage{pbox}
\usepackage{color}
\usepackage{array}
\newcolumntype{L}[1]{>{\raggedright\let\newline\\\arraybackslash\hspace{0pt}}m{#1}}

\newcommand{\R}{\mathbb{R}}
\DeclareMathOperator*{\argmin}{arg\,min}

\definecolor{mypink}{RGB}{219, 48, 122}

\newcommand{\vertbars}[1]{\lVert #1 \rVert}

\title{The Typology of Polysemy: A Multilingual Distributional Framework}

\author{
Ella Rabinovich$^\dagger$ \hspace{3cm}
Yang Xu$^{\dagger*}$ \hspace{3cm}
Suzanne Stevenson$^\dagger$ 
\vspace{0.2cm} \\
$^\dagger$Department of Computer Science, $^*$Cognitive Science Program
\vspace{0.1cm} \\
University of Toronto \vspace{0.1cm} \\
\texttt{\{ella,yangxu,suzanne\}@cs.toronto.edu}
}

\begin{document}

\maketitle

\begin{abstract}
Lexical semantic typology has identified important cross-linguistic generalizations about the variation and commonalities in polysemy patterns---how languages package up meanings into words.  Recent computational research has enabled investigation of lexical semantics at a much larger scale, but little work has explored lexical typology across semantic domains, nor the factors that influence cross-linguistic similarities. We present a novel computational framework that quantifies \textit{semantic affinity}, the cross-linguistic similarity of lexical semantics for a concept. Our approach defines a common multilingual semantic space that enables a direct comparison of the lexical expression of concepts across languages. We validate our framework against empirical findings on lexical semantic typology at both the concept and domain levels. Our results reveal an intricate interaction between semantic domains and extra-linguistic factors, beyond language phylogeny, that co-shape the typology of polysemy across languages.


\end{abstract}

\begin{figure}[bt]
\centering
\includegraphics[width=9.25cm, trim = 3.25cm 5.5cm 5cm 2.5cm, clip]{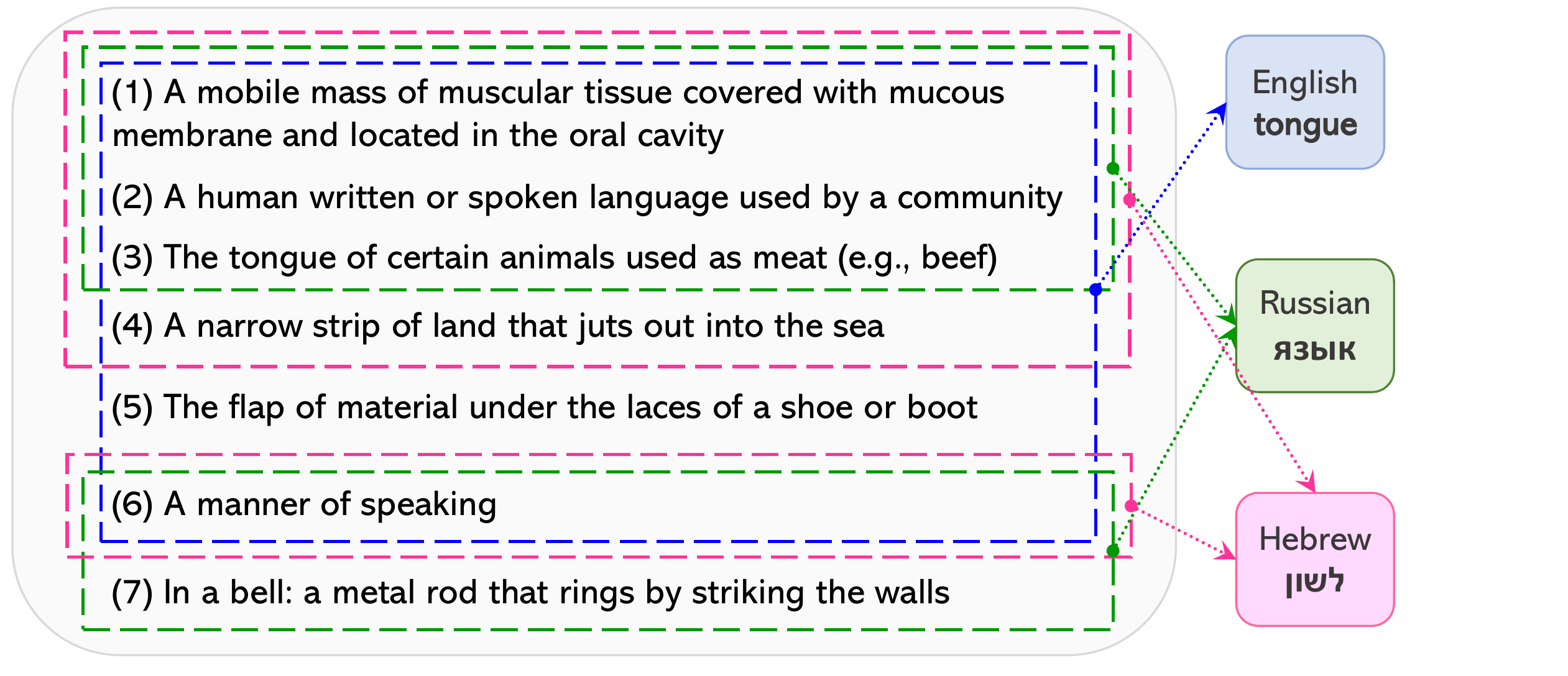}
\vspace{-0.6cm}
\caption{A partial list of meanings given in Babelnet for the English word ``tongue'', as well as for the corresponding Russian and Hebrew word forms. 
}
\label{fig:babelnet-senses}
\end{figure}

A central issue in cognitive science is the nature of the mental mapping between language and the world.  One oft-studied question is how and why languages vary in the way they use words to partition semantic space (e.g., \citeNP{berlin69,levinson03}).
Polysemy---the use of a single word form to express multiple related senses---is a fundamental property of language that exemplifies this variation.
%
Figure~\ref{fig:babelnet-senses} shows how 
 word forms across languages may differ in the sets of senses they cover. 
Despite this variation, there is also much cross-linguistic commonality in word meanings, as seen in the overlap of sense expression in Figure~\ref{fig:babelnet-senses}.
How much do languages vary in their lexical semantics, and what contributes to the observed cross-linguistic patterns of polysemy? Here we present a principled and large-scale computational approach to these questions. 

Work in typology---studies of the constrained variation exhibited by languages---has 
identified important cross-linguistic generalizations regarding polysemy patterns across many semantic domains.
For example, some research has focused on the primitives that underlie cross-linguistic lexical categorization (e.g., \citeNP{berlin69, levinson03}), while other work has studied the degree of universality of polysemy patterns (e.g., \citeNP{majid2015semantic, youn2016universal}).
However, such studies have been restricted in scope due to reliance on manual methods.
To find robust answers to the above questions on semantic typology, automatic methods are required to enable larger scale study.


Computational research has proposed various methodologies to explore lexical semantic structure at a more comprehensive scale. Previous work exploiting distributional representations has studied how language-pair semantic similarity is influenced by various factors, including phylogeny 
(e.g., \cite{thompson2018quantifying, beinborn2019semantic}), geography \cite{eger2016language}, culture \cite{thompson2018quantifying}, and conceptual structure \cite{xu20}. To the best of our knowledge, the only study considering lexical variation at the level of semantic domain is that by \citeA{thompson2018quantifying}. That study used monolingual word embeddings for quantifying cross-linguistic semantic alignment in an inherently multilingual setting. Importantly, previous work has typically focused on descriptive analyses that are not evaluated against the empirical generalizations 
reported in the literature. 

We propose a novel framework\footnote{All code and data are available at \url{https://github.com/ellarabi/semantic-affinity}} for quantifying lexical semantic variation that addresses these limitations in two respects. First, we develop a measure of \textit{semantic affinity} that assesses the degree of semantic similarity of the corresponding word forms for a concept across many languages.  We take an alternative approach to existing work in which we construct a \textit{common multilingual semantic space} that enables a direct comparison of the lexical expression of concepts across multiple languages.  Second, we evaluate our approach against known findings in the typological literature to assess the validity of our measure.


\section{Background on Lexical Semantic Variation}
\label{sec:related-work}


We focus on some key findings regarding lexical variation and factors that influence it, at the level of individual concepts, domains, and languages.  We summarize these empirical findings in Table~\ref{findings}, which we will assess our framework against.

\vspace{-0.2cm}
\paragraph{Individual Concepts.} 
Semantic change is an important source of polysemy, and factors that influence that process may also influence the degree of semantic affinity of concepts.  A number of studies have shown that the rate of semantic change is correlated with the psycholinguistc factors of frequency and degree of polysemy (estimated by number of senses), and minimally correlated with word length (a proxy of frequency) when frequency and polysemy are both controlled for (e.g., 
\citeNP{hamilton2016diachronic}).
\citeA{pagel2007frequency} also found that numbers (e.g., ``two'') are slowest to change among the grammatical categories, which follow a specified order (Table~\ref{findings}).

\vspace{-0.2cm}
\paragraph{Semantic Domains.} Recently, much research on lexical semantic typology has studied cross-linguistic universals and principles of variation in patterns of polysemy (e.g., \citeNP{berlin69,levinson03,majid2015semantic,youn2016universal}). Two studies in particular enable us to assess the relative level of semantic affinity across different semantic domains. First, \citeA{majid2015semantic}, 
using naming tasks to elicit lexical data, manually determine an ordering of the degree of semantic variation among four conceptual domains across 12 Germanic languages; see Table~\ref{findings}.  Second, \citeA{youn2016universal}, using manual translation across $81$ languages, found that a set of $23$ basic concepts pertaining to the physical environment exhibits ``universal tendencies'' in lexical semantics---i.e., has a high degree of cross-linguistic similarity. In particular, they show that this similarity is unaccounted for by phylogeny, geography, or climate (with one exception, which we return to in our results). Interestingly, \citeA{regier2016languages} did find an effect of environmental factors on the cross-linguistic lexicalization of ``snow'' and ``ice'' (but this subdomain is too small to assess). 

\vspace{-0.2cm}
\paragraph{Language-Level Influences.} 
Studies quantifying similarity between pairs of languages have exploited distributional properties extracted from monolingual (e.g., \citeNP{beinborn2019semantic, thompson2018quantifying})
or bilingual \cite{eger2016language} semantic spaces. The findings by and large highlight the correlation between languages' semantic and \textit{phylogenetic} similarity 
\cite{beinborn2019semantic}. Correlations of geographical \cite{eger2016language} and cultural \cite{thompson2018quantifying} factors with cross-linguistic semantic similarity have been shown.  However, an analysis of their influence across various semantic domains, and evaluation against empirical observations, have been lacking.

\begin{table*}[]
\center
\begin{tabular}{l | l}
Level of analysis & Summary of empirical findings from the literature  \\ \hline \hline
Individual &  Rate of semantic change correlates with frequency (-), polysemy (+), word length ($\approx$0) \\
concepts &  Lexical evolution rate: number$<$pro.$<$adv.$<$noun$<$verb$<$adj.$<$conj.$<$prep. \\ \hline
Semantic & Semantic variation: Color, Body Parts$<$Containers$<$Spatial Relations  \\
domains & Universal tendencies in lexicalizing basic concepts of the physical environment \\ \hline
Language-level & Language phylogeny correlates with semantic similarity across languages\\
influences & Environmental factors (geography/climate) influence lexical semantic typology \\
\end{tabular}
\vspace{-0.1cm}
\caption{Condensed summary of recent findings on lexical semantic typology to which we compare our approach.}
\label{findings}
\end{table*}

\vspace{-0.2cm}
\paragraph{Our Approach.}  Our work is closely related to that by \citeA{thompson2018quantifying}, who presented a large-scale study of cross-linguistic semantic alignment at the level of the domain.  That study used monolingual semantic spaces to quantify cross-linguistic semantic alignment, where the similarity between words representing a concept in two languages was estimated indirectly through the proximity of these words to their (partial) neighbourhood in individual spaces. Our work explores an alternative approach based on \textit{semantic affinity}, which differs in that we: (1)~quantify cross-linguistic semantic similarity in a direct and unmediated way, by constructing a common multilingual semantic space shared across languages of interest; (2)~evaluate this framework against empirical findings in the literature, a critical aspect that was not explored in the previous work; and (3)~leverage this framework to perform analysis of factors---both linguistic and extra-linguistic---that influence semantic affinity of a concept, at the levels of a single concept and a domain. 

\section{Datasets}
\label{sec:datasets}

\paragraph{Translation Sets.} Measuring cross-linguistic semantic affinity of a concept requires a set of words representing that concept in various languages. We used NorthEuralex \cite{dellert2017northeuralex}, a large lexicostatistical database providing accurate (manual) translations of over $1000$ basic concepts in $107$ languages from $20$ distinct language families of Northern Eurasia, including over $30$ Indo-European languages. Each concept, represented by a corresponding German term, is annotated for part-of-speech (POS), and links to a set of word forms representing this concept in other languages. This yields a set of common concepts, spanning multiple domains, and including accurate translations of the same concept into words across multiple languages.  Since these words naturally have various additional meanings across the languages, reflecting various patterns of polysemy, this introduces a natural testbed for our analysis. Despite limitations due to known quality control issues,\footnote{See note at \url{http://northeuralex.org/}.} this is one of the most comprehensive multilingual datasets, suitable for this study.

\paragraph{Cross-Linguistic Polysemy Data.} BabelNet \cite{navigli2010babelnet} is a very large multilingual semantic network in which each node represents a language-independent meaning, to which words across the represented languages can link.  For example, as illustrated in Figure~\ref{fig:babelnet-senses}, the node for the meaning ``A human written or spoken language used by a community'' will be linked from the English word ``tongue'', as well as the corresponding words in Russian and Hebrew. Crucially, as seen in the figure, our target words that represent the same NorthEuralex concept in different languages may cluster different (sub)sets of meanings -- sharing a common set of meanings, but deviating in language-specific ones. For each of our concepts, we document the total number of \textit{distinct} meanings associated with it cross-linguistically, as accessed through the words representing this concept in the set of languages used in this work. We restricted the list of concepts to those supported in Babelnet by at least $30$ of the $35$ languages considered in this study (i.e., at least $30$ of languages have a corresponding word-form entity in the database).  This results in $697$ concepts across many domains.


\section{Computational Framework}
\label{sec:methods}

Our goal is to measure the degree of semantic similarity of the corresponding words for a concept across many languages. We adopt a distributional semantics approach given the success of such models in capturing subtleties of word meaning (e.g., \citeNP{hollis2016principals, pereira2016comparative}).
We construct \textit{multilingual} common semantic spaces that enable the projection of words from multiple languages into a shared space (e.g., \citeNP{conneau2017word}).
Specifically, words in different languages that have roughly the same meaning are brought close to each other within a single vector space.
For a given concept, we operationalize its semantic affinity across languages by the degree of similarity of the corresponding words' representations in the common semantic space. This notion of affinity can be extended to a semantic domain (a collection of concepts) and to languages (across all concepts).



\vspace{-0.2cm}
\paragraph{Building a Multilingual Semantic Space.} We use the Facebook MUSE framework \cite{conneau2017word}, shown to obtain good results on many tasks (e.g., \citeNP{artetxe2019massively, beinborn2019semantic}), for construction of a multilingual semantic space. The model uses a set of automatically extracted bilingual dictionaries between pairs of languages to project monolingual word representations in two languages onto a common space.  It does so while optimizing the mutual proximity of representations of an automatically extracted set of translation equivalents (words referring to the same entity; e.g., English ``apple'', French ``pomme''). Using English as a pivot language, the procedure can then be scaled to any number of languages L, assuming an English--L bilingual dictionary, and ultimately resulting in a common massively multilingual semantic space. Further details on this procedure can be found in Appendix A.

\begin{figure}[hbt]
\centering
\resizebox{\columnwidth}{!}{
\includegraphics{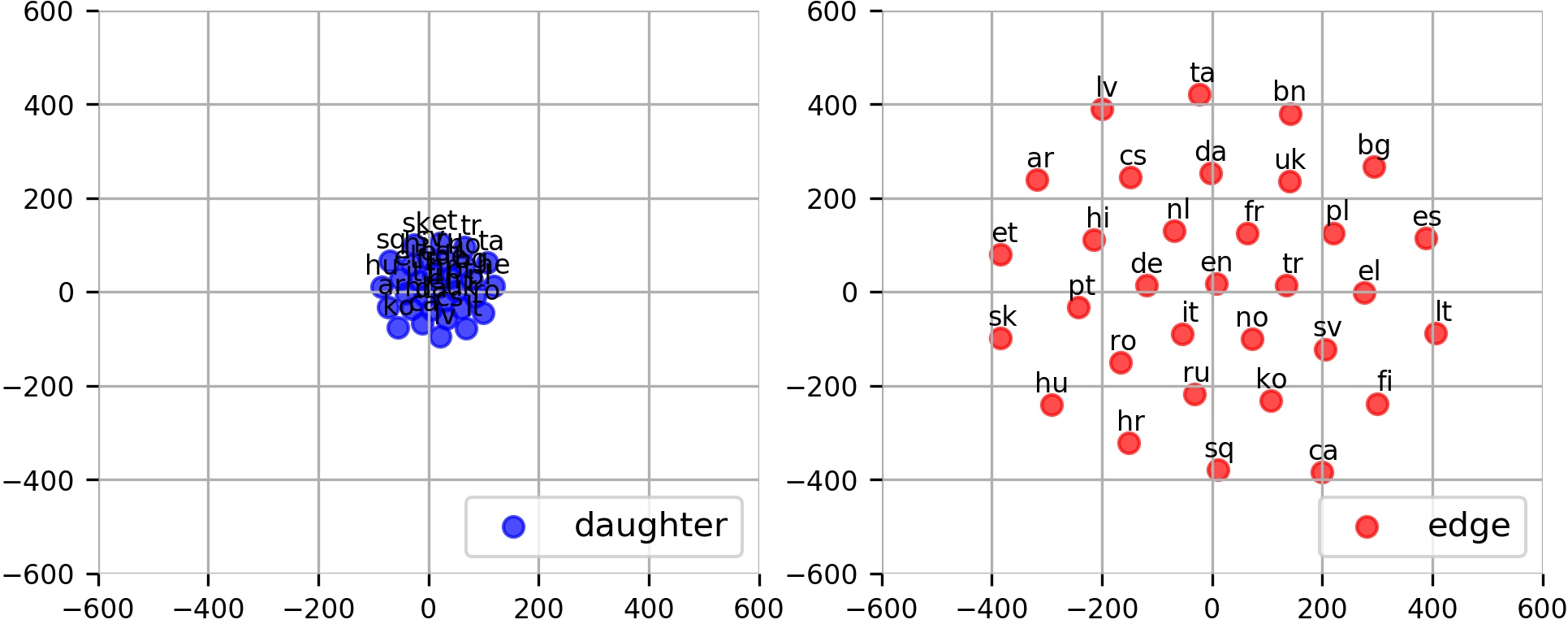}
}
\caption{t-SNE projections of multilingual embeddings, corresponding to English terms ``daughter'' and ``edge''.}
\label{fig:tsne-embeddings}
\end{figure}

For building our multilingual space, we use the set of $35$ geographically-diverse languages supported by NorthEuralex, Babelnet and MUSE bilingual dictionaries, and the corresponding fastText monolingual embeddings \cite{grave2018learning}. In training and validation, we excluded the entire set of words representing our target concepts from the list of translation equivalents whose proximity is optimized by MUSE in creating the common embedding space. Figure~\ref{fig:tsne-embeddings} illustrates that different concepts can have differing degrees of cross-linguistic similarity in the resulting common semantic space. 

\vspace{-0.2cm}
\paragraph{Quantifying Semantic Affinity.} The semantic affinity of a concept w.r.t.\ a set of languages amounts to the mutual proximity of embeddings representing the concept across various languages; that is, the ``tighter'' the cluster of embeddings, the more (cross-linguistically) similar the concept is (cf.\ Figure~\ref{fig:tsne-embeddings}). Formally, given a concept $c$ and a set of $N$ word forms representing $c$ across a set of languages $\mathcal{L}{=}\{L_1, L_2, ..., L_N\}$, we denote its corresponding 300-dimensional vector representations by $V^c=\{v^c_1, v^c_2, ..., v^c_N\}$. 
Using cosine similarity as our similarity metric, we compute the centroid of $V^c$ by calculating the average of its 
constituents. This procedure results in a vector in the direction of the cluster centroid:

\begin{equation}
Cent(V^c) = \frac {1}{N} \sum\limits_{i} v^c_i, \quad i{\in}[1..N], \vertbars{v^c_i}{=}1
\label{eq:centroid}
\end{equation}

We then estimate cross-linguistic semantic affinity of $c$ with respect to $\mathcal{L}$ by computing its \textit{cluster density}, specifically, we average over individual words' cosine similarities to the (virtual) cluster centroid in Equation \ref{eq:centroid} ($i\in[1..N]$):

\begin{equation}
SemAff(V^c) = \frac {1}{N} \sum\limits_{i} cos(v^c_i, Cent(V^c))
\label{eq:density}
\end{equation}

Intuitively, semantic affinity mirrors the extent of meaning similarity of a concept as expressed across a set of languages.  For example, as expected from Figure~\ref{fig:tsne-embeddings}, we find higher semantic affinity for the concept corresponding to ``daughter'' ($0.766$) than for that corresponding to ``edge'' ($0.572$).

\section{Results on Concepts and Domains}

We first evaluate how well our measure of cross-linguistic semantic affinity matches empirical findings at the level of individual concepts and semantic domains (Table~\ref{findings}).

\subsection{Semantic Affinity of Concepts}
\label{sec:predictors}

We hypothesize that factors that play a role in lexical semantic change (within a language) may also influence the degree of semantic affinity across languages. We thus suggest the following variables as predictors of cross-linguistic affinity:

\noindent
\textbf{Mean Word Rank.} 
We derive a ranked list of the top-N words in each language using frequencies recorded in wordfreq\footnote{\url{https://pypi.org/project/wordfreq/}} 
For a given concept $c$, we then average the ranks of its corresponding words across the languages.\footnote{Word frequency (which strictly correlates with rank in a language) is incomparable across different languages.}

\noindent
\textbf{Degree of Polysemy.}  We computed the total number of unique senses of the words associated with a concept across our languages (see the Datasets section for details).

\noindent
\textbf{Mean Word Length.} We computed the average length of word forms corresponding to a concept across our languages.

We perform multiple regression analysis using the semantic affinity of concepts (\mbox{SemAff}, Equation~\ref{eq:density}) as the dependent variable, and the predictors above as independent variables; see Table~\ref{tbl:word-affinity-regression}. All variables together explain nearly 40\% (adj.\ $r^2{=}{0.381}$) of the variance.
Our results are in line with previous findings on the relation of these psycholinguistic variables to semantic change (cf.\ Table~\ref{findings}), as we expected since lexical evolution is an important source of polysemy.  Mean word rank is negatively correlated with semantic affinity, implying that less frequently used concepts have lower cross-linguistic semantic affinity. As well, concepts with a higher degree of polysemy exhibit higher cross-linguistic semantic diversity. 
Finally, mean word length shows the weakest correlation with affinity among the variables.

\begin{table}[h]
\centering
\begin{tabular}{lrrrr}
predictor & coef.($\beta{\times}10$) & std err($\beta$) & t-stat  \\ \hline
coeff & 6.615 & 0.001 & 445.747  \\
mean word rank & -0.242 & 0.002 & -13.294  \\
degree of polysemy & -0.200 & 0.002 & -16.037  \\
mean word length & 0.129 & 0.001 & 8.640  \\
\end{tabular}
\vspace{-.1in}
\caption{Multiple regression analysis. The response variable is concept semantic affinity, a real value in the $0{-}1$ range. $p<.001$ in all cases.}
\label{tbl:word-affinity-regression}
\end{table}

We further computed cross-linguistic semantic affinity for various POS categories; that is, by averaging SemAff over concepts that share the same POS in the NorthEuralex dataset, requiring a minimum of five concepts per tag. Table~\ref{tbl:domain-affinity} (left) reports the results.  Here too we find that our relative rankings replicate the relative stability over time of these categories as found by \citeA{pagel2007frequency}, with a single exception of a swap in ordering between verbs and nouns (cf.\ Table~\ref{findings}).

\begin{table*}[hbt]
\centering
\begin{tabular}{lrrr||lrrr}
domain & \multicolumn{1}{c}{count} & \multicolumn{1}{c}{SemAff} & \multicolumn{1}{c||}{SD} & \multicolumn{1}{c}{POS} & \multicolumn{1}{c}{count} & \multicolumn{1}{c}{SemAff} & \multicolumn{1}{c}{SD} \\ \hline
NUMERAL & 21 & 0.701 & 0.034 & Numerals & 21 & 0.701 &  0.034 \\
ADVERB & 37 & 0.672 & 0.050 & Youn et al. Set & 22 & 0.683 & 0.031 \\
VERB & 204 & 0.668 & 0.041 & Colors & 9 & 0.675 & 0.028 \\
NOUN & 474 & 0.656 & 0.052 & Body Parts & 42 & 0.643 & 0.033 \\
ADJECTIVE & 102 & 0.645 & 0.038 & Spatial Relations & 8 & 0.621 & 0.043 \\
PREPOSITION & 5 & 0.631 & 0.046 & Containers & 9 & 0.611 & 0.030 \\
\end{tabular}
\vspace{-.06in}
\caption{Cross-linguistic semantic affinity and standard deviation by part-of-speech and domain.}
\label{tbl:domain-affinity}
\end{table*}

\begin{figure}[h]
\centering
\includegraphics[width=9cm, trim = 1.65cm 5cm 1cm 3.5cm, clip]{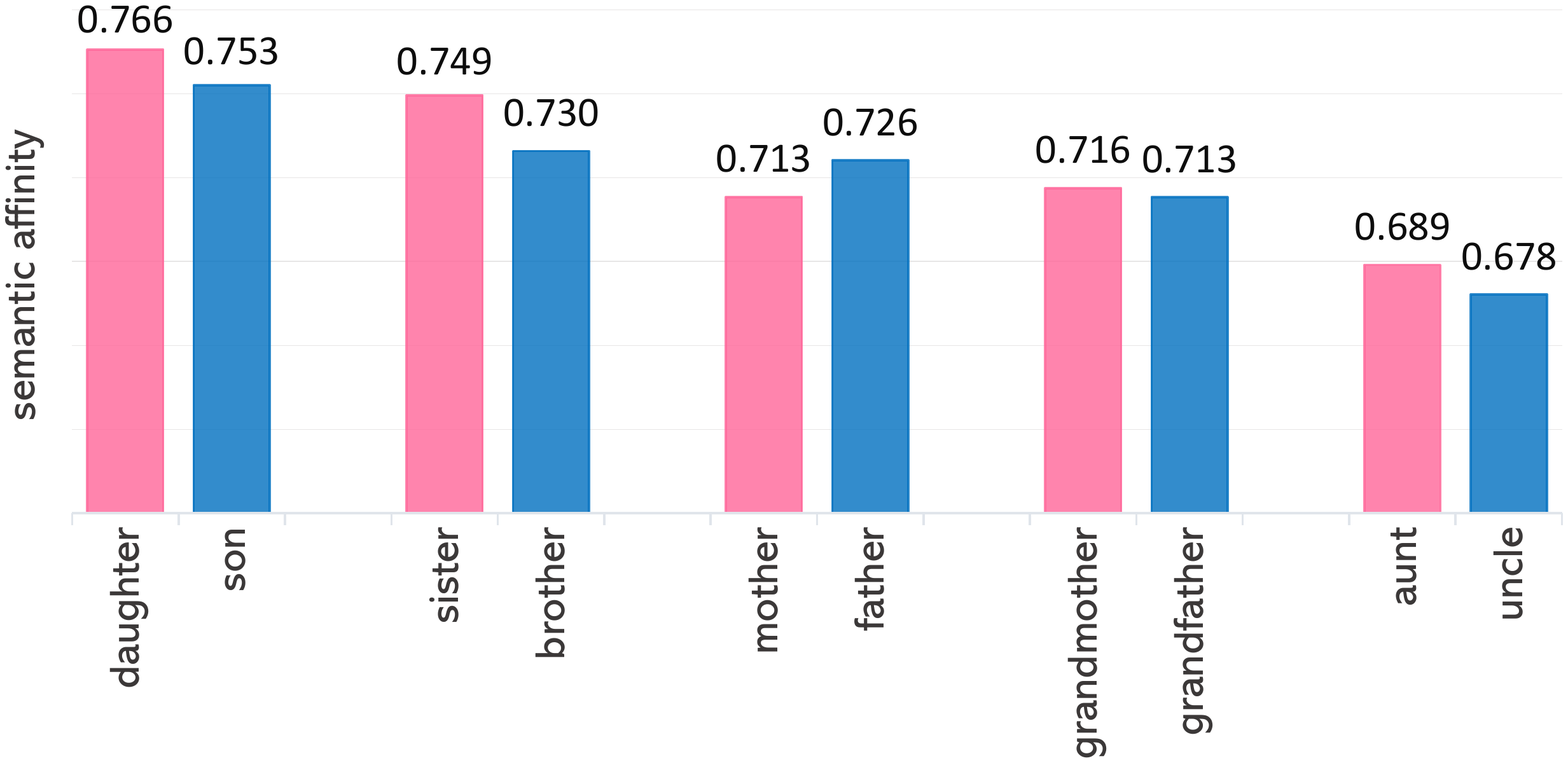}
\vspace{-0.25in}
\caption{Semantic affinity of kinship female and male terms located by their relatedness to ego (left-to-right).} \label{fig:kinship}
\end{figure}

To provide a fine-grained qualitative view of our framework, we visualize semantic affinities of 10 common concepts in the domain of kinship. Figure~\ref{fig:kinship} reveals an interesting symmetry in this domain. Specifically, semantic affinity is higher for kin terms that are more closely related to ego than those farther away, and this trend is symmetric between male and female kin types. Concretely, ``aunt'' and ``uncle'' (and ``grandmother'' and ``grandfather'') show the lowest semantic affinity across languages, in comparison to the closer kin relations such as children, siblings, and parents of ego. This observation is consistent with independent empirical findings suggesting that remote kin terms, e.g., ``aunt'' and ``uncle'', are most often extended to unrelated persons \cite{ballweg}.

\subsection{Semantic Affinity across Domains}

We derive cross-linguistic semantic affinity at the level of a domain by averaging the SemAff of its individual concepts. We use semantic domains similar\footnote{Restricted by the set of words used in this study, and by limited availability of data used in previous work.
} to those used by \citeA{majid2015semantic} and \citeA{pagel2007frequency}, as well as the set of $23$ concepts analysed by \citeA{youn2016universal}, denoting this of words set as the ``Youn et al. Set'' hereafter.

Table~\ref{tbl:domain-affinity} (right) reports the SemAff of each domain along with its number of concepts. The results generally support fundamental observations in the literature: specifically, that words used for Numerals, Colors, and Body Parts have greater semantic affinity across languages than Spatial Relations (relational words) and Containers (artifacts). Our approach thereby provides additional empirical evidence for theoretically-motivated hypotheses on the nature of lexical semantic structure across semantic domains. 
However, our findings do not strictly mirror the results reported by \citeA{majid2015semantic}, e.g., spatial relations and containers are ranked in the opposite order; that, possibly due to the slightly different set of concepts in both categories, and the much larger set of languages used in our experiment. The relatively high affinity of the Youn et al. Set reflects the empirical finding by \citeA{youn2016universal} regarding the \textit{universal semantic structure} of the underlying set of words---a finding that our results suggest does not systematically generalize to additional domains.

\section{Results on Language-level Influences}
\label{sec:lang-pair-analysis}

Above we considered semantic affinity of concepts and domains; we can also calculate a measure of semantic affinity between languages (across concepts from a range of domains). As noted earlier, there is much evidence that such broad semantic affinity between languages is correlated with phylogenetic similarity, but the evidence is sparser and less clear regarding the influence of other factors, such as geography and climate.
Here, we extend both these strands of work, by considering the influence of phylogeny, geography, and/or climate on a large scale sample of concepts and domains across languages. We expect genealogical similarities between languages to be predictive of their semantic affinity.  Moreover, we hypothesize that geography and climate exhibit predictive power on semantic similarity above and beyond genealogy, thereby highlighting the effect of environmental factors on shaping lexical semantic systems.

\subsection{Semantic Distance Between Languages}

We can measure semantic affinity between two languages w.r.t.\ a single concept as the cosine similarity between the projection of the two words representing that concept onto our common semantic space. We then define semantic affinity between two languages across a set of concepts as the average such similarity across the individual concepts. Finally, to align with terminology in the literature on phylogenetic \textit{distance} (as opposed to similarity), we convert this semantic affinity measure to a semantic distance by subtracting it from 1. Then, given a set of concepts $\mathcal{C}$, semantic distance (SDist) between two languages $L_i$ and $L_j$ w.r.t.\ $\mathcal{C}$ is
defined as:

\vspace{-0.2cm}
\begin{equation}
SDist(L_i,L_j) = 1 - \frac {1}{|\mathcal{C}|}\sum cos(v^c_i, v^c_j), \quad c{\in}\mathcal{C}
\label{eq:pairwise}
\end{equation}

We limit the following analysis to $22$ IE languages in our set,\footnote{We excluded English and Spanish because their widespread native use prevents isolating their geographic and climate data.} which have well-established historical data.

\subsection{Phylogenetic and Environmental Factors}

We use a well-accepted tree \cite{gray2003language} for computing phylogenetic distances between pairs of languages. 
We define phylogenetic distance between two languages as their (unweighted) path length in the tree.
We further model two environmental factors: geographical and climate distances between languages.
We model language-pairwise geographical distance as the Euclidean distance between their corresponding (longitude, latitude) coordinates in the NorthEuralex database.  We model language-pairwise climate distance as the Euclidean distance between their climate vectors.  These vectors are formed from \textit{temperature} and \textit{precipitation} data in a climate database \cite{kottek2006world}.  For each language, we extract this data from the region whose (longitude, latitude) coordinates are closest to those of the language in NorthEuralex.

\subsection{Language-level Results and Discussion}

Pairwise correlations show that the three predictor variables---phylogeny (PHY), geography (GEO), and climate (CLM)---are correlated with each other as expected: a high correlation between PHY and GEO (Pearson's $r=0.559$; genetically related languages are often close in space), and between GEO and CLM ($0.807$; regions close in space generally have similar climates). Importantly, all three predictors also exhibit a significant association with semantic distance (SDist): $0.402$ for PHY, $0.518$ for GEO and $0.516$ for CLI. Notably, a higher correlation of SDist is found with geographical and climate predictors than with phylogenetic distance, suggesting that these have a considerable effect on lexical semantic structure.

We next perform a multiple regression analysis to estimate the relative contribution of each of the three factors to predicting language-pair semantic distance. The three independent variables together explain $30\%$ of SDist variance (adj.\ $r^2{=}{0.301}$); Table~\ref{tbl:language-pairwise-regression} reports the details. Although all predictors share similar coefficients, the highest coefficient is assigned to climate distance, implying its substantial predictive power on semantic diversity of concepts in our data. The contribution of geographical distance appears only marginally significant, likely due to its interaction with climate.

\begin{table}[hbt]
\centering
\resizebox{7.75cm}{!}{
\begin{tabular}{lrrrr}
 & coeff. & std err &   \\ 
predictor & \small{($\beta{\times}10$)} & \multicolumn{1}{c}{\small{($\beta$)}} & t-stat & pval \\ \hline
coeff & 5.589 & 0.003 & 202.396 & \textless{0.001} \\
PHY & 0.086 & 0.003 & 2.587 & 0.005 \\
GEO & 0.087 & 0.006 & 1.452 & 0.063 \\
CLM & 0.141 & 0.006 & 2.505 & 0.006 \\
\end{tabular}
}
\vspace{-.05in}
\caption{Multiple regression analysis predicting SDist.}
\label{tbl:language-pairwise-regression}
\end{table}

\vspace{-0.2in}
\paragraph{Evidence beyond Language Phylogeny.}

\begin{figure*}[h!]
\centering
\includegraphics[width=\textwidth]{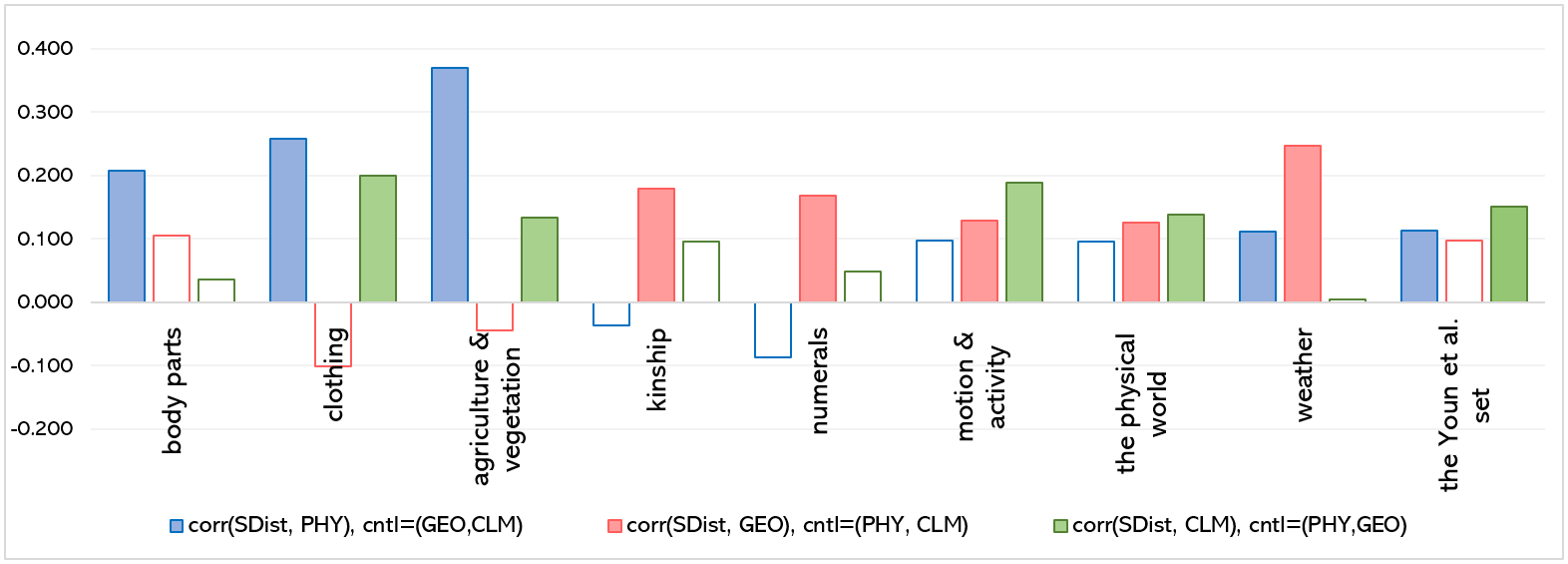}
\caption{Partial correlation tests across focused domains. Each bar shows the correlation of language-pair SDist with one of the factors (PHY, GEO, and CLM) while controlling for the other two. Empty bars show non-sig. correlation ($p$\textgreater$0.05$).}
\label{fig:domain-word-affinity}
\end{figure*}

We hypothesize that the effects of environmental factors (GEO and CLM) are not uniformly distributed across domains: we expect them to be  most evident in concepts that are more subject to interpretation associated with environmental conditions. We construct focused sets of concepts that represent a variety of semantic domains presumed to be affected by environmental factors to a varying extent. For example, the Clothing domain includes words like ``shirt'', ``cap'' and ``boot''; Motion\&Activity includes words like ``ski'', ``boat'', ``sway''; and the Weather domain includes words like ``cloud'', ``frost'' and ``thunder''. We perform partial correlation tests of the predictive power of each of the predictors---PHY, GEO and CLM---on pairwise language distance within each domain (while controlling for the other two).  Figure~\ref{fig:domain-word-affinity} presents the results.

Our methodology reveals an intricate interaction between the semantic domains and the influencing factors. We found that each factor---not just phylogeny---plays a non-trivial role in explaining the cross-language semantic affinities, and our results are in accord with some independent findings from the literature. In particular, phylogeny is the strongest and the only significant predictor in the domain of Body Parts. This finding is consistent with evidence that semantic shifts in body-part terms provide important clues to proto-language reconstruction \cite{ matisoff1985out}. In contrast, we observed that phylogeny alone is not sufficient to account for affinity of Clothing, where climate would naturally co-shape the semantic typology.  Additionally, geography is a salient factor in the domains of Kinship and Numerals, which relates to findings that suggest kinship networks vary along geographical areas \cite{murphy2008variations}, and that numeral systems in a language family are shaped by areal diffusion \cite{epps2005areal}. Finally, the significant effect of climate on the Youn et al. Set---a domain that is reported to exhibit cross-linguistic ``universals''---mirrors their own finding that the split of languages by \textit{humid} and \textit{arid} areas was an exceptional case to their universal semantics hypothesis (cf.~\citeNP{youn2016universal}).

\section{Conclusions}

Lexical semantic typology reflects both variation and commonalities in the patterns of polysemy across languages. We proposed a principled and large-scale approach to the study of cross-linguistic lexical semantic structure at the levels of individual concept and semantic domain. We evaluated our framework against existing findings in previous studies, demonstrating results that conform to established fundamentals pertaining to semantic variation across languages. Through the analysis of a subset of Indo-European languages, our framework discovered that extra-linguistic factors of geography and climate carry over explanatory value regarding semantic variation between languages---above and beyond genealogical relations. Our work suggests that the environment may play an important role in explaining the cross-linguistic variation in polysemy.

Despite these advances, our approach is currently limited by the reliance on a manually curated dataset to provide the translation equivalents across languages of the concepts we investigate.  In the future, we plan to apply our framework on translation equivalents extracted automatically (e.g., via word alignment in bilingual corpora), thereby extending it to additional concepts and languages.

\section{Acknowledgements}

ER and SS are funded through NSERC grant RGPIN-2017-06506 to SS.
YX is funded through an
NSERC Discovery Grant, a SSHRC Insight Grant, and a Connaught New Researcher
Award.

\bibliographystyle{apacite}
\setlength{\bibleftmargin}{.125in}
\setlength{\bibindent}{-\bibleftmargin}

\bibliography{main}

\section*{Appendix A}
\label{sec:appendices}

\label{sec:appendixA}
In this appendix we lay out some intuition regarding the construction of a multilingual semantic space. The procedure involves fixing a pivot language (e.g., English), and performing multiple steps of alignment of two semantic spaces (e.g., English and French), thereby generating a \textit{bilingual} space. The process can be further scaled up to an arbitrary number of languages, pairwise aligned with the pivot, and, therefore, with each other. The essence of the construction of a bilingual semantic space lies in aligning two monolingual spaces. The input to the alignment process includes an (automatically or manually constructed) dictionary of $n$ words in two languages $\{x_i,y_i\}, i{\in}\{1,..,n\}$, and two matrices---$X$ and $Y$---containing $d$-dimensional representations (embeddings) of the $n$ words in the two languages: source (represented by the matrix $X$) and target (represented by $Y$). The alignment procedure then learns a linear mapping (matrix $W^*$) between the source and the target semantic space such that:

\begin{equation}
W^* = \argmin_{W{\in}M_{d}(\R)} ||WX-Y||_{F},
\label{eq:alignment}
\end{equation}

\noindent
where $d$ is the dimension of the embeddings, $M_{d}(\R)$ is the space of $d{\times}d$ matrices of real numbers, and $X$ and $Y$ are two matrices of size $d{\times}n$ containing the embeddings of the words in the aligned vocabulary \cite{conneau2017word}. The `$F$' notation on the right-hand side of Equation~\ref{eq:alignment} denotes extracting the matrix norm (a single number) by applying the Frobenius norm~\cite{datta2010numerical}, defined as:

\begin{equation}
||A||_{F} = \sqrt{\sum_{i=1}^{m} \sum_{j=1}^{n} |a_{ij}|^{2}}
\label{eq:frobenius}
\end{equation}

\noindent for an arbitrary matrix A with $m{\times}n$ dimensions.

Polysemous extensions are preserved in the bilingual dictionaries by mapping a single word form with multiple meanings in a certain language into distinct words in another language, i.e., $m{\times}n$ mapping. As such, the French word `mandat' is mapped into two English translation equivalents: `mandate', `warrant' in the automatically extracted  French-English dictionary. All binlingual dictionaries used in this work are those provided by~\citeA{conneau2017word}.

\end{document}